\pdfoutput=1
\documentclass[sigconf, nonacm]{acmart}

\usepackage{booktabs}
\usepackage{graphicx}
\usepackage{amsmath}
\usepackage{etoolbox}
\usepackage{subcaption}
\usepackage{float}
\usepackage{algorithm}
\usepackage{algpseudocode}

\definecolor{headblue}{HTML}{1F3A68}
\DeclareRobustCommand\eqcontribmark{%
  \textcolor{headblue}{\scalebox{1.25}{$\bigstar$}}}
\newcommand\preprintmark{%
  \textcolor{headblue}{\sffamily\bfseries\footnotesize PREPRINT}}
\newcommand\awardnote{In press. This paper received the Best Paper Award at
  BAMS-ORSI 2024, the International Conference on Emerging Trends in Business
  Analytics \& Management Sciences and 57th Annual Convention of the
  Operational Research Society of India (ORSI-2024).}
\makeatletter
\patchcmd{\@mktitle@i}{\@title}{\textcolor{headblue}{\@title}}{}{}
\patchcmd{\@mktitle@iii}{\@title}{\textcolor{headblue}{\@title}}{}{}
\patchcmd{\@mktitle@iv}{\@title}{\textcolor{headblue}{\@title}}{}{}
\preto\@secfont{\color{headblue}}
\preto\@subsecfont{\color{headblue}}
\preto\@subsubsecfont{\color{headblue}}
\let\preprint@fnsymbol\@fnsymbol
\def\@fnsymbol#1{\ifnum#1=\@ne\eqcontribmark\else\preprint@fnsymbol{#1}\fi}
\AtBeginDocument{%
  \let\preprint@standard\ps@standardpagestyle
  \def\ps@standardpagestyle{\preprint@standard
    \fancyhead[RO,LE]{\preprintmark}}%
  \let\preprint@first\ps@firstpagestyle
  \def\ps@firstpagestyle{\preprint@first
    \fancyhead[RO,LE]{\preprintmark}}%
  \pagestyle{standardpagestyle}}
\makeatother

\begin{document}
\title[Financial Context-Rich DRL Strategy for Trading]{Financial
  Context-Rich Strategy for Actor-Critic based Agents: A Deep Reinforcement
  Learning Approach for Stock Trading}

\author{Hrishikesh Singh}
\authornote{Both authors contributed equally to this research.}
\orcid{0009-0003-2496-4947}
\affiliation{%
  \institution{Department of Applied Mathematics \& Scientific Computing,
    Indian Institute of Technology Roorkee}
  \city{Roorkee}
  \country{India}
}

\author{Aarti Sharma}
\authornotemark[1]
\orcid{0009-0003-7181-1257}
\affiliation{%
  \institution{Department of Applied Mathematics \& Scientific Computing,
    Indian Institute of Technology Roorkee}
  \city{Roorkee}
  \country{India}
}

\author{Millie Pant}
\orcid{0000-0002-7668-7887}
\affiliation{%
  \institution{Department of Applied Mathematics \& Scientific Computing,
    Indian Institute of Technology Roorkee}
  \city{Roorkee}
  \country{India}
}
\email{pant.milli@as.iitr.ac.in}

\renewcommand{\shortauthors}{Singh, Sharma and Pant}

\begin{abstract}
    In modern financial markets, more than ${{3}/{4}}^{th}$ of the large volume trade is executed by algorithmic trading based on advanced trading strategies to attain higher precision and speed than traditional human traders. The conventional strategies focused on price-based indicators are inadequate in capturing the market dynamism arising from real-world events such as social sentiments, political instability, or sudden economic shifts. With the rise of \emph{Deep Reinforcement Learning} (DRL) and data availability, there is a requirement for more efficient algorithms that can incorporate these external factors in trading strategies. DRL has shown potential in simulating real-world scenarios and in simplifying complex decision-making since DRL agents can automate the trading process \emph{i.e.} executing trades in a manner comparable to human traders. They have shown reliable performance since they can update their policies based on risk and quantitative factors. In this paper, we propose a DRL approach based on partial and deep-infusion of sentimental analysis. It features distinct policy designs of \emph{actor-critic-based} agents: only policy optimization (\emph{A2C} \& \emph{PPO}) and policy optimization along with Q-learning principles (\emph{DDPG, TD3} \& \emph{SAC}).  Our findings reveal, that the agents based on policy optimization combined with Q-learning, significantly and consistently demonstrate improved performance upon sentiment deep-infusion in profit maximization as well as risk-averse trading.
\end{abstract}



\maketitle
\begingroup
  \renewcommand\thefootnote{}%
  \footnotetext{\awardnote}%
\endgroup


\section{Introduction}
{
    In 2024, India will remain the world's fastest-growing major economy. According to the World Bank’s GDP ranking 2024 list India is the 5th largest economy, inevitably with rapid development, an increase in social media usage, the spread of the digital economy, and financial literacy more and more retail investors are investing their money in stock markets. Hence, the need to develop optimal trading strategies that are profitable and minimize risk is paramount. The stock market is a dynamic and rapidly changing environment, and many traditional trading strategies based on stochastic calculus, numerical analysis, and quadratic programming\cite{HARRISON1983313} may prove to be suboptimal since they fall short in utilizing all the dynamically evolving relevant factors that influence the equities such as stocks.

    Deep learning-based methods show potential in this scenario since they are capable of extracting relevant features from huge datasets. Supervised learning-based approaches to predict stock prices mainly follow two steps, that is, forecasting and decision-making\cite{Pahwastock2019}. But such strategies may lead to suboptimal performance due to “\emph{forecasting bottleneck}”. Reinforcement Learning (RL) solves this problem since it is based on “\emph{learn by experience}” methodology, whereas supervised learning is based on “\emph{learn by example}”. It shows excellent performance in stochastic environments such as the stock market due to its ability to make sequential decisions\cite{Jae2001stock,WU2023120474}.

    However, the bottleneck of reinforcement learning approaches is its inefficiency in scaling with the huge amount of information processing required for real-world scenarios. In this scenario, deep reinforcement learning fits as a solution. Deep reinforcement learning (DRL) uses agents or robots to develop trading strategies in complex scenarios involving numerous features. The agent mainly works by either optimizing its policy or the value function. The agents that we have used are mainly those that optimize policy and those that optimize policy and value function.

    The stock price movement is highly volatile and is not solely based on quantitative features such as the performance metrics of a company and the technical indicators. Many important scenarios that impact the country’s fate such as elections, budget, etc. have seen significant movement in stock prices. Thus, considering quantitative features only in stochastic decision-making processes is painting only half a picture, it is necessary to include qualitative features as well such as the sentiment value.

    The main aim of our research is to build and validate automated trading strategies using actor-critic DRL agents like A2C, PPO, TD3, SAC, DDPG by incorporating sentiment analysis  and thus improve decision-making capabilities. In our paper, we have incorporated the sentiment value in the representational state space as well as in the reward calculations to see their impact on influencing the agent’s behavior in developing trading strategies. Our main contributions are as follows:

    \begin{itemize}
        \item A comprehensive evaluation of distinct DRL agents in financial market conditions.
        \item Integration of sentiment and analyzing its impact on the trading performance of agents.
        \item Examining the impact of policy design variations on financial decision-making.
    \end{itemize}
}


\section{Related Work}
{
    Stock market prediction has been of interest to researchers for an extended time. The earlier studies utilized linear\cite{Cakra2015StockPP} and machine learning techniques\cite{Pai2005497} to build predictive models. Over time various evolved techniques such as artificial neural networks\cite{TICKNOR20135501} were employed to predict stock prices, options prices, exchange rates, etc. Some techniques involved extracting qualitative information using text mining to predict the stock prices\cite{Oliveira2017125,Hagenau2013685}. Several works used natural language processing techniques to forecast financial instrument movement\cite{Yildirim20184719}.

    In the deep learning landscape, the authors of \cite{Cao2019stockprice} proposed a modified CNN and time series analysis of financial data to build a forecasting model for stock prices. In a similar vein, in \cite{SEZER2018525} CNN is used on stock chart images to develop a financial trading model. Recurrent neural networks show excellent performance on time series data. They have also been employed in predictive stock trading, the authors of \cite{HuynhStock2017} used a Bidirectional GRU network and compared its performance with LSTM and GRU. Attention-based LSTM was another method proposed for stock prediction \cite{inproceedings}.LSTM with optimization algorithms is explored by the author of \cite{GULMEZ2023120346}. Reinforcement learning has an advantage in simulating real-life trading scenarios and faster and superior decision-making. There has been ongoing research in this direction establishing the effectiveness of RL-based models for stock prediction\cite{Jae2001stock, WU2023120474}. Deep reinforcement learning combines the feature extraction capabilities of deep learning with real-world simulation and complex decision-making abilities of reinforcement learning. DRL-based models are capable of providing useful insights as well as planning and developing complex trading strategies\cite{ZOU2024122801}.

    The existing work establishes the effectiveness of deep learning-based agents. Building upon this foundation we have further explored and compared the decision-making abilities of five distinct DRL agents which have unique policy designs, in the same market conditions. We further experimented with sentiment infusion, to analyze the trading behavior of these agents upon providing more context of the financial market.
}

\section{Dataset}
{
    \subsection{Stock Market Data}
    For the stock historical dataset, we selected 5 top Indian stocks by market capitalization: Reliance, Adani, Bharti Airtel, HDFC and ICICI, covering main financial sectors such as energy, telecommunication, banking. The main advantages of selecting stocks by market capitalization are, stocks with larger market capitalization tend to be more stable and less volatile compared to smaller-cap stocks.

    \begin{table}[!h]
        \centering
        \renewcommand{\arraystretch}{1.2} 
        \begin{tabular}{|c|c|c|}
        \hline
        Company          & Market Capitalization      & Sector               \\
                         & (in Trillion INR)          &                      \\ \hline
        Reliance         & 19.64 T                    & Oil                  \\ \hline
        Adani Ports      & 3.09 T                     & Infrastructure/Ports \\ \hline
        Bharti Airtel    & 9.44 T                     & Telecommunications   \\ \hline
        HDFC Bank        & 12.53 T                    & Banking              \\ \hline
        ICICI Bank       & 8.71 T                     & Banking              \\ \hline
        \end{tabular}
        \vspace*{+7pt}
        \caption{Market Capitalization \& Domain of Selected Indian Companies (Sept 2024)}\label{tab:market-cap}
    \end{table}

    Larger companies often have more comprehensive and accurate financial data available. This can be advantageous for RL algorithms, as high-quality data is crucial for training models and making informed investment decisions. Large-cap stocks typically have a significant influence on the overall market. By focusing on these stocks, RL models can align more closely with broader market trends and indices.

    The range of our financial dataset is 5.5 years from Jan 2019 to June 2024. We obtained the historical data of the selected stocks from Yahoo Finance. The features of historical stock data from Yahoo Finance are: \emph{Date, Ticker, Open, High, Low, Close, Volume} \& \emph{Day}. We have also pre-processed out financial data by adding technical indicators, \emph{CBOE Volatility Index} (VIX) and turbulence.

    The technical indicators are: \emph{Moving Average Convergence Divergence} (MACD), \emph{Bollinger Bands, Relative Strength Index} (RSI-30), \emph{Commodity Channel Index} (CCI-30), \emph{Directional Index} (DX-30), \emph{30-day Simple Moving Average} (Close 30 SMA), \emph{60-day Simple Moving Average} (Close 60 SMA).

    \subsection{Financial News}
    We build the corresponding financial news dataset for the prior selected stocks using Google’s RSS Feeds API. To obtain a broad context and reporting for the stocks we collected multiple news for the same stock at any given day among the timeline we had decided earlier (~5 years).

    The multiple headlines for a particular day help us obtain a more realistic average market sentiment for that day. On an average for a particular day the number of headlines vary from 1 to 10. The features of our financial news dataset are: \emph{Date, Ticker, News Headline} and \emph{Source}.

    \subsection{Sentiment Score}
    After scraping the headlines, we did a sentiment analysis of these headlines using FinBERT, (BERT for Financial Text Mining).

    The authors of FinBERT\cite{10.5555/3491440.3492062}, have used two financial corpuses to further pre-train the BERT model. They used TRC2-financial which consists of 46,143 documents with more than 29M words and nearly 400K sentences. The second dataset is Financial PhraseBank\cite{malo2013gooddebtbaddebt} which consists of 4845 English sentences selected randomly from financial news found on LexisNexis database.

    FinBERT generates 3 normalized label scores for each headline i.e. positive, negative and neutral where ‘positive’ label is encoded as 1, ‘negative’ label as -1 and ‘neutral’ label as 0.  We assigned the label with the highest score to corresponding news.

    For a single day we had ~up to 10 headlines. For each headline we calculated the sentiment label and sentiment score. On the days when the news headlines were not available through Google’s RSS Feed API, we assigned a sentiment score of 0 and sentiment label as neutral.

    Our final data with news headlines and sentiment analysis was of dimension: (4843, 4) with standard features: Date, Ticker, Sentiment Label and Sentiment Score.

    From the sentiment score we obtained for each date we calculated an inclination score in the following way:

    Let s be a particular stock, d be a particular date. Let $\eta(s,d)$ denote the news headline of a particular stock s on a date d. Since we have gathered multiple news headlines for a stock s on a particular day, d, we define $\eta_i(s,d)$ be the $i^{th}$ news headline for stocks s on date d, where $i$ ranges from 1 to N, where N denotes the total number of articles on a particular date for a particular stock.  For each $\eta_i(s,d)$ we multiplied the sentiment score with \{1, -1 or 0\} depending on the sentiment label to obtain an adjusted sentiment score ranging from -1 to 1. Then we took the mean of the adjusted sentiment score to obtain an inclination score. More formally:

    \begin{equation}
        \theta = \frac{\sum_{i=1}^{N}\alpha(\eta_i(s,d))}{N}
        \label{equation:confidence-score}
    \end{equation}

    $\theta$: Inclination score for news headline \emph{i} for stock \emph{s} and date \emph{d}

    $\alpha$: Adjusted Sentiment Score for news headline i for stock s and date d

    $N$: Total number of news headlines for a stock s on date d

    \subsection{Fusion}
    After obtaining the inclination score for each date for a stock we merged these scores with the historical stock data. We merged these two datasets on stock’s ticker symbol and date. The missing values of inclination score for a particular stock and date in the merged dataset were populated with a default value of 0 , thereby indicating a neutral sentiment.
}


\section{Modeling}
{
    To model this research problem, we have replicated the trading activity in market structure onto agent and environment based framework. The trading activity is primarily a \emph{Markov Decision Process} (MDP), where an agent is an actor, taking the action by interacting with the dynamic environment.

    Every trading activity by the DRL model is a series of Markov decisions. The Environment consists of the states, reward and metadata. The sequential decisions taken by the agents over a course of action are represented as a list of states and the associated rewards with each step. By continuous feedback from the dynamically changing environment, agents learn to trade better by policy optimization, which is a component of agents.

    \subsection{Stock Market Structure}{
        US stock market have two major exchange i.e. New York Stock Exchange (NYSE) and NASDAQ. Similarly, Indian stock market also have dual-exchange market structure i.e. National Stock Exchange (NSE) and Bombay Stock Exchange (BSE). \\

        These exchanges handle various kinds of instruments as a part of their day-to-day trading. We will be focusing on stock trading among these instruments.

        \begin{figure}[!htbp]
            \centering
            \includegraphics[width=\linewidth,
                alt={Schematic of the National Stock Exchange (NSE) architecture. External entities (members, regulators, data vendors, custodians, clearing banks, depositories) connect through leased lines, a switch and the internet to the online systems: order book and trading, risk management, and clearing and settlement, which feed member data, surveillance and a data warehouse. Market data feeds, indices and a feed handler carry market data out to external entities.}]{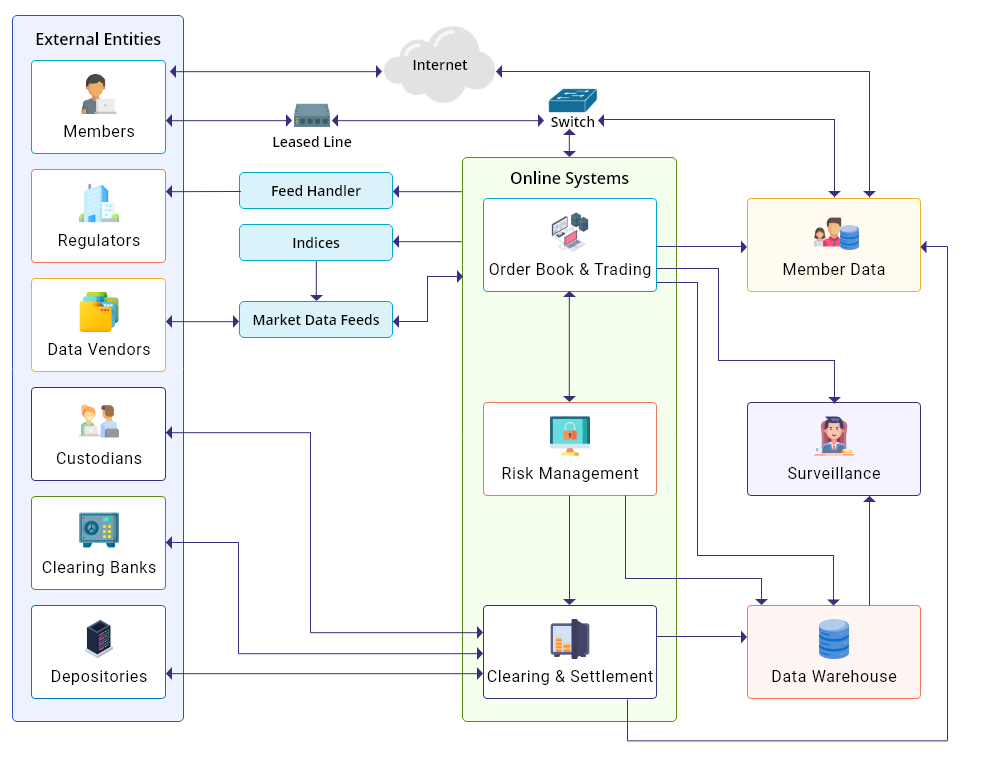} 
            \caption{NSE Schema}
            \label{fig:NSE}
        \end{figure}

        In a live trading scenario, there are two components working together i.e. Trader (Human) and the Stock Trading Setup (Orderbook, Market Datafeed, Capital, etc.). A trader will analyze the market using the inbound data feed to strategize their action for achieving their objective. There are various aspects to the objective such as risk appetite, capital, desired liquidity, stakeholder preference, etc. so here the strategy is influenced by the financial experience of the trader or portfolio manager combined with the derived knowledge from the analysis of the incoming feed.

        Our research will be localized between Order Book \& Trading and Market Data Feeds in the architecture diagram.

        To model and simulate this real world setup we have replicated the trading activity and a component of stock market architecture relevant to our use case in the form of four major components: \emph{Environment, State, Reward} and \emph{Agent}.

        \begin{figure*}[t]
            \centering
            \includegraphics[width=0.8\textwidth,
                alt={Reinforcement learning architecture for trading. The agent (an actor-critic neural network with experience replay, learning rate and an exploration mechanism) acts on the environment (market conditions). The environment calculates the state (price data, technical indicators, portfolio information, sentiment data) and the reward (portfolio value, risk metrics, profit and loss, transaction cost, sentiment data), which are fed back to update the agent and its policy (deterministic or stochastic policy, policy gradient, exploration versus exploitation).}]{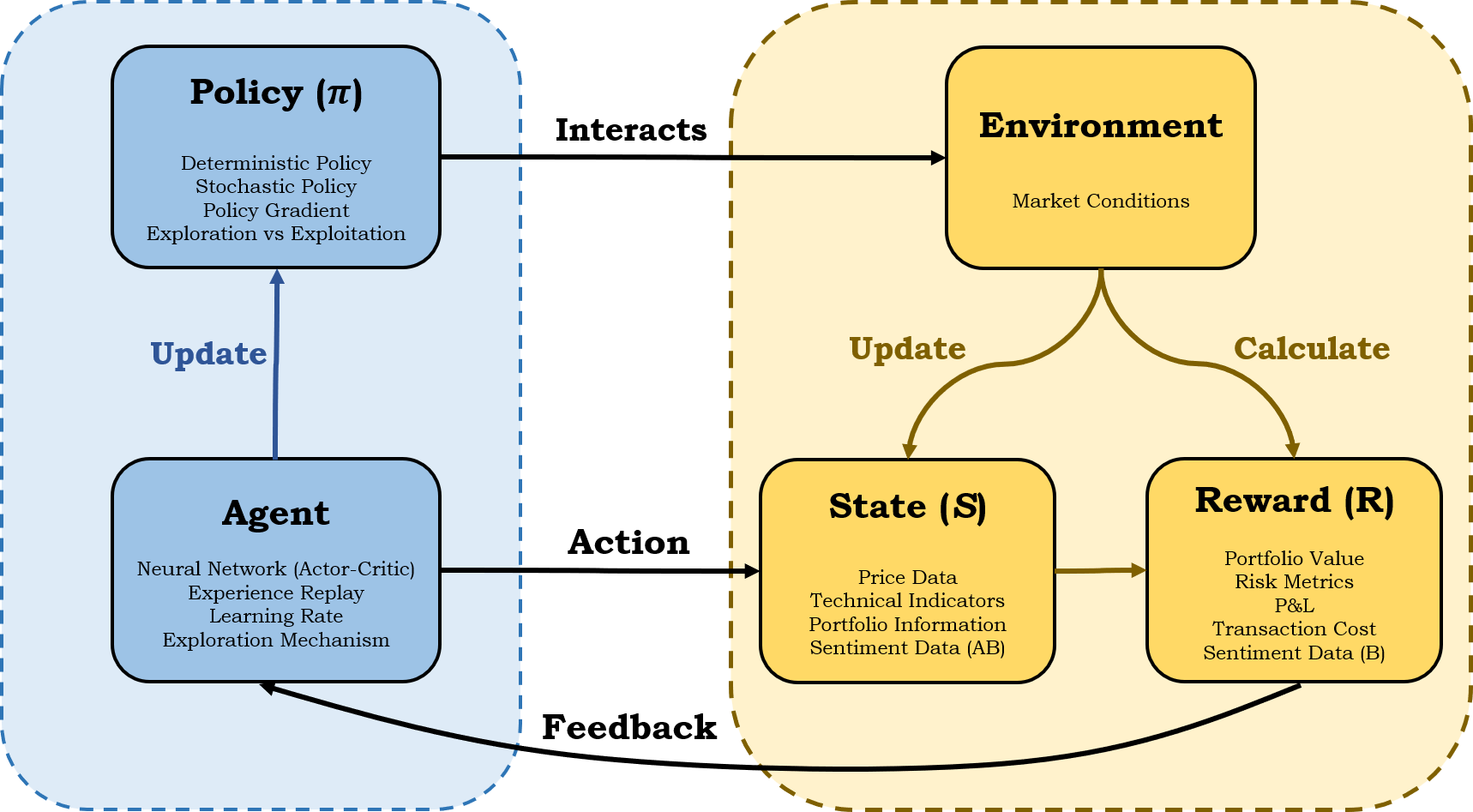} 
            \caption{Trading based Reinforcement Learning Architecture}
            \label{fig:model}
        \end{figure*}

        For a simpler analogy, the stock market architecture and data such as the exchange in which we are trading, rules of engagement, market conditions will be a part of the modeled environment. The current scenario or snapshot of our position in the market at a specific time will be called \emph{State} i.e. price, portfolio composition, values of technical indicators etc. Now the \emph{Agent} assumes the role of a trader and has similar options available as to a human trader. While \emph{Policy} is analogous to the intuition or technical acumen of a trader whereas \emph{Reward} can be assumed as the feedback for the actions taken by the model.

        Now we will explain the components of our modeling in following sections:

    }

    \subsection{State}{
        A state describes the current conditions of an environment to an agent. Based on these conditions an RL agent takes actions. States in reinforcement learning follow Markov property which means that given the current state the future must be independent of the past. We have modeled our use case as a \emph{Markov Decision Process}. The key aspect of MDP in RL is that decisions and rewards are assumed to be dependent on the current state only. Hence, the state must contain all the information about the current conditions of the environment. We have defined a finite state space. In our case we have included these features in our state space: agent’s cash balance, the current stock prices, number of shares held for each stock, technical indicators, and inclination score derived from the sentiment score and sentiment label for each stock. The need to include sentiment data in state space is to let the agent take more sentiment aware decisions.  Since we have defined a Markov state we can define the transition dynamics of our environment at time step \emph{t} as:

        \begin{equation}
            P(\zeta_{t+1}, \alpha_{t+1}|\zeta_t, \alpha_t)
            \label{equation:state}
        \end{equation}

        Here \emph{s} and \emph{t} are state and action respectively at current time step \emph{t} and the next time step. \emph{t+1}. The equation \ref{equation:state} describes the probability of our environment to transition from a state $\zeta_t$ and action $\alpha_{t}$ to the next state $\zeta_{t+1}$ and action $\alpha_{t+1}$. The probability to transition from current state and action to the next state and action is entirely dependent upon current state and action.
    }

    \subsection{Environment}{
        In reinforcement learning an environment encompasses state, action and reward. Essentially, it is the environment with which an RL agent interacts. Based on the state the agent takes actions and rewards or penalties are provided by the environment for the action taken. The transition dynamics for the environment as in equation ~\ref{equation:state} gives the probability distribution of transitioning to the next state given the current state and action. In our use case equation ~\ref{equation:state}, \ref{equation:reward} and \ref{equation:reward-step4} define various aspects of our environment.
    }

    \subsection{Reward}\label{sec:reward}{
        There are two ways we have calculated reward. In the case of stock trading without sentiment, as well as the case where sentiment incorporation is done only in the state, the reward is calculated as follows:

        Let us assume we are at time step $t$. At the beginning of each time step, we calculate the asset value as $\beta(t_s)$, which is the sum of cash balance $(c)$ and the number of stocks held $(n)$. After the agent performs certain actions, the asset value at the end of the time step is calculated similarly as $\beta(t_e)$. The final reward $(\gamma)$ is obtained as the change in the asset value from the beginning of the time step to the end of the time step as follows:

            \begin{equation}
            \gamma = \Delta(\beta(t)) = \beta(t_e) - \beta(t_s)
            \label{equation:reward}
            \end{equation}

        \vspace*{+10pt}
        In another case, where we have incorporated the inclination score in the reward calculation so that sentiment has a better influence on the agent’s actions, we have followed this strategy:

        As in equation ~\ref{equation:reward} above, first the asset change is calculated. Then we compute the average individual reward. The calculation of the average individual reward is done as follows:

        \textbf{Step 1:} For a particular stock, calculate the difference between the current price $\rho_{t_{i}}$ and the price at the beginning of the time step $\rho_{t_{i-1}}$:
            \begin{equation}
            \Delta \rho_t = \rho_{t_{i}} - \rho_{t_{i-1}}
            \label{equation:reward-step1}
            \end{equation}

        \textbf{Step 2:} Calculate the individual reward $r_s$ for a particular stock $(s)$ by multiplying the price difference with the inclination score $\theta_i$:
            \begin{equation}
            r_s = \Delta (\rho_t) \times \theta_i
            \label{equation:reward-step2}
            \end{equation}

        After calculating the individual reward for each stock, we take the mean of all individual rewards to obtain the average individual reward:
            \begin{equation}
            r_{\text{avg}} = \frac{\sum_{s}^{} r_s}{N}, \quad \text{where} \ N \ \text{is the number of stocks}
            \label{equation:reward-step3}
            \end{equation}

        The final reward is then calculated as:
            \begin{equation}
            \gamma = \Delta (\beta(t)) + r_{\text{avg}}
            \label{equation:reward-step4}
            \end{equation}

        In both these cases, we have obtained various results, which are documented in the experiment section.
    }

    \subsection{Agent}{
        In the case of stock trading, there are three actions available to the agent i.e. buy, sell or hold. These actions result in creation of the next state and based on the performance, a reward or penalty is given to the agent.

        \begin{figure}[!htbp]
            \centering
            \includegraphics[width=0.85\linewidth,
                alt={Markov decision chain of agent actions. From state S1 the agent can hold and stay in S1, buy and move to state S2, or sell and move to state S3; from S2 and S3 it can again buy, sell or hold.}]{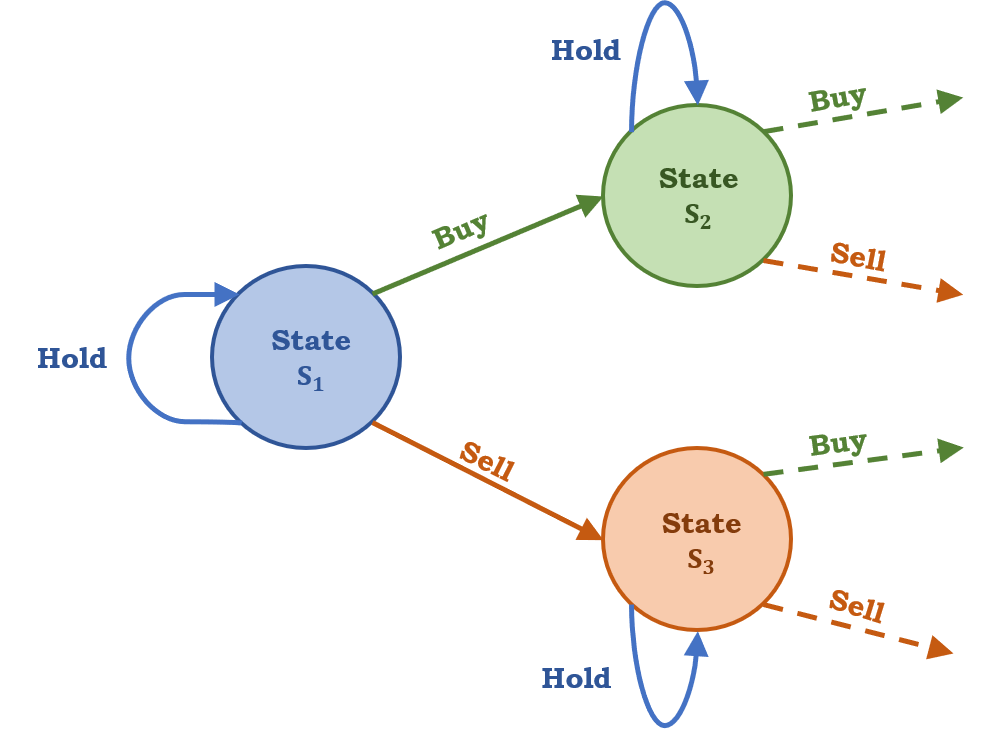}
            \caption{Markov Decision Chain of Agent actions}
            \label{fig:agent}
        \end{figure}

        Using this reward and state feedback, the agent aims to optimize its policy to attain the objective faster and accurately. Policy can be optimized by actor-critic methods directly or by using Q-learning in continuous action spaces.

        We have selected 5 advanced reinforcement learning agents (Actor-Critic) to balance the exploration and exploitation in our modeling environment. A2C and PPO are based on policy optimization while DDPG, SAC and TD3 extend to use Q-learning too. The characteristics and the workings are defined as follows:

        \begin{figure}[!htbp]
            \centering
            \includegraphics[width=0.75\linewidth,
                alt={Venn diagram of the selected algorithms. Policy optimization methods (A2C, PPO, TRPO) and Q-learning methods (DQN, C51, QR-DQN, HER) overlap in DDPG, TD3 and SAC. A2C and PPO from policy optimization and DDPG, TD3 and SAC from the overlap are highlighted as the five agents used.}]{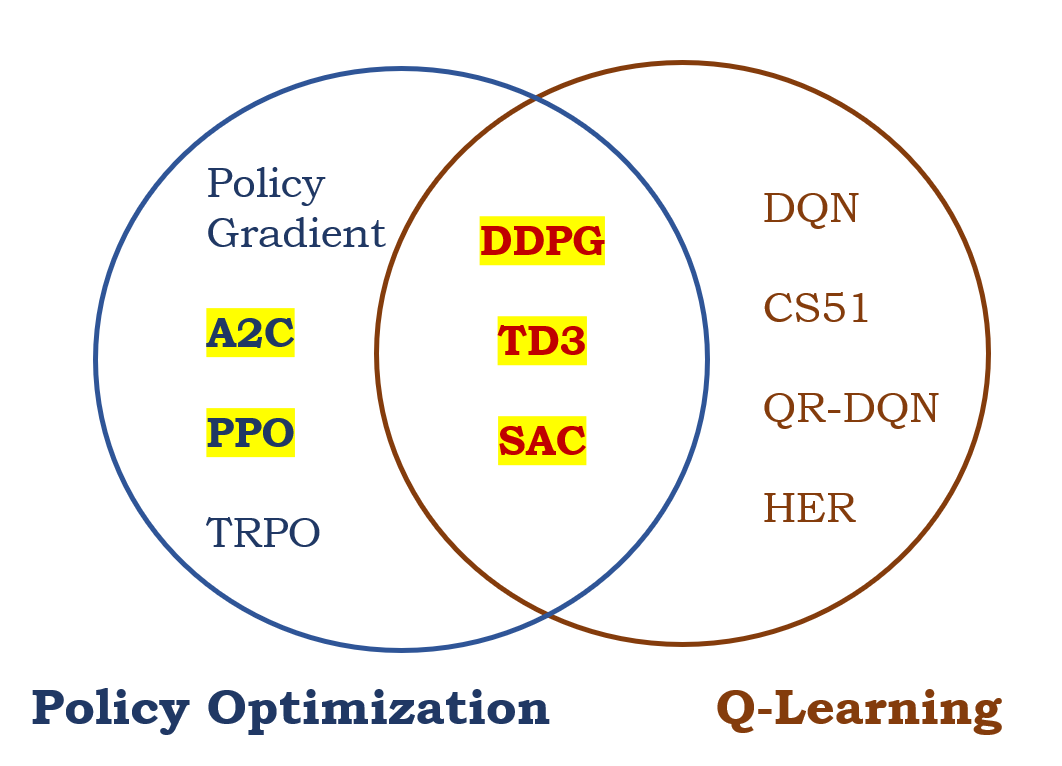}
            \caption{Selected Algorithms}
            \label{fig:algo}
        \end{figure}

        \subsubsection{Advantage Actor-Critic (A2C)}{
            The actor-critic method is based on policy gradient and primarily has two neural networks, an actor and a critic. The actor determines the next action while getting post-action feedback about their impact from the critic network. This step is simplified further by employing an advantage function which acts as an estimator by minimizing variance and giving faster learning speed. The A2C is synchronous while A3C is asynchronous\cite{haarnoja2018softactorcriticoffpolicymaximumm}. In A3C the parallel update step reduces the inter-dependence of actor-learner for one to finish before starting the other process.

            In our trading experiment, A2C intends to maximize profits by improving its trading policy based on actor actions and feedback from the critic.

            \vspace{+5pt}
            \textbf{Policy Gradient Update:}

            \begin{equation}
            \nabla_{\omega} U(\omega) = {E}_{x \sim p^{\mu}} \left[\nabla_{b}  R^{\mu}(x, b) \big|_{b=\mu(x)} \nabla_{\omega} \mu_{\omega}(x)\right]
            \label{equation:a2c}
            \end{equation}

            Where $\nabla_{\omega} U(\omega)$ represents the gradient of the expected return $U$ with respect to policy parameters $\omega$.
            $R^{\mu}(x, b)$ is the action-value function, estimating the expected return for taking action $b$ in state $x$ and then following policy $\mu$.
            $\mu_{\omega}(x)$ represents the policy\cite{haarnoja2018softactorcriticoffpolicymaximumm}, a mapping from states to actions, parameterized by $\omega$.

            \vspace{+5pt}
            \textbf{Advantage Function:}

            \begin{equation}
            A(x, b) = R^{\mu}(x, b) - S^{\mu}(x)
            \label{equation:a2c-advantage}
            \end{equation}

            Where $A(x, b)$ is the advantage of taking action $b$ in state $x$.
            $S^{\mu}(x)$ is the state-value function\cite{haarnoja2018softactorcriticoffpolicymaximumm}, estimating the expected return of starting in state $x$ and following policy $\mu$.
        }

        \subsubsection{Proximal Policy Optimization (PPO)}{
            This is an on-policy algorithm which minimizes rapid policy changes to improve stability by using clipped probability ratios in its objective function.

            We have selected this agent algorithm\cite{schulman2017proximalpolicyoptimizationalgorithms} to avoid making significant trade volume especially when market volatility is high. The clipped surrogate probability function equips the algorithm to evaluate the continuous action space and hence improves the stability and mitigating the arbitrary trade risks.

            \vspace{+5pt}
            \textbf{Clipped Surrogate Objective:}

            \begin{equation}
            \rho_i(\theta) = \frac{\pi_{\theta}(b_i|x_i)}{\pi_{\theta_{old}}(b_i|x_i)}
            \label{equation:ppo-objective}
            \end{equation}

            Where $\rho_i(\theta)$ is the probability ratio of taking action $b_i$ in state $x_i$ under the current policy $\pi_{\theta}$ compared to the old policy $\pi_{\theta_{old}}$.\cite{schulman2017proximalpolicyoptimizationalgorithms}

            This ratio is then clipped:

            \begin{equation}
            \text{clip}(\rho_i(\theta), 1-\epsilon, 1+\epsilon)
            \label{equation:ppo-ratio}
            \end{equation}

            Where $\epsilon$ is a hyperparameter limiting the policy update step size.
        }

        \subsubsection{Deep Deterministic Policy Gradient (DDPG)}{
            This is an off-policy algorithm based on an actor-critic architecture but utilizes deterministic policy gradient theorem\cite{schulman2017proximalpolicyoptimizationalgorithms}. Here the advantage is generated by using a replay buffer which can store and sample experiences requiring less data and smoother learning. Instead of providing an action probability distribution, it simply suggests the exact action to take, hence the name “deterministic”. This algorithm also works on continuous action space\cite{lillicrap2019continuouscontroldeepreinforcement,haarnoja2018softactorcriticoffpolicymaximumm}.

            \vspace{+5pt}
            \textbf{Actor Update:}

            \begin{equation}
            \nabla_{\alpha_{\nu}}  U \approx {E}_{x \sim \rho_{\beta}} \left[\nabla_{b} R(x, b | \alpha_R) \big|_{b=\nu(x|\alpha_{\nu})} \nabla_{\alpha_{\nu}} \nu(x | \alpha_{\nu}) \right]
            \label{equation:ddpg-actor}
            \end{equation}

            Where $\nabla_{\alpha_{\nu}}  U$ is the gradient of the expected return $U$ with respect to the actor\cite{fujimoto2018addressingfunctionapproximationerror} parameters $\alpha_{\nu}$.
            $R(x, b | \alpha_R)$ represents the Q-value, the expected return of taking action $b$ in state $x$, parameterized by $\alpha_R$.
            $\nu(x | \alpha_{\nu})$ represents the deterministic policy, directly mapping states to actions, parameterized by $\alpha_{\nu}$.

            \vspace{+5pt}
            \textbf{Critic Update:}

            \begin{equation}
            R^{\mu}(x, b) =  {E} \left[r(x, b) + \gamma R^{\mu}(x', b')\right]
            \label{equation:ddpg-critic}
            \end{equation}

            Where $x'$ is the next state. $b'$ is the action taken in the next state following the current policy.
        }

        \subsubsection{Twin Delayed DDPG (TD3)}{
            In continuous action spaces, one of the problems is overestimation bias in Q-learning. To solve this problem, an enhancement in the DDPG algorithm led to TD3\cite{fujimoto2018addressingfunctionapproximationerror}. This algorithm proposed 3 additions: target policy smoothing, delayed policy update and clipped double Q-learning.

            In our experiment, TD3 minimizes the trading risk originating from overestimation bias and also reduces the variance. To attain this objective, it uses 2 Q-networks and provides a conservative action value by adopting a minimum of all predictions hence lowering the risk even if the market volatility is unusually high because of noisy data.

            \vspace{+5pt}
            \textbf{Target Value Update:}

            \begin{equation}
            z = r + \gamma \min_{i=1,2}  R_{\alpha'_i}(x', \nu_{\beta'}(x') + \zeta)
            \label{equation:td3-update1}
            \end{equation}

            \begin{equation}
            \zeta \sim \text{clip}(\mathcal{N}(0, \sigma), -c, c)
            \label{equation:td3-update2}
            \end{equation}

            Where $z$ is the target value.
            $R_{\alpha'_i}(x', a)$ represents the target Q-values from two separate critics ($i=1,2$).
            $\nu_{\beta'}(x')$ is the target policy.
            $\zeta$ is the clipped noise term, sampled from a clipped normal distribution, added to the target action\cite{fujimoto2018addressingfunctionapproximationerror}.
        }

        \subsubsection{Soft Actor-Critic (SAC)}{
            SAC is an off-policy algorithm based on the actor-critic framework but with the concept of maximum entropy reinforcement learning\cite{haarnoja2018softactorcriticoffpolicymaximumm}. The conventional RL prioritizes maximizing the expected return while SAC tweaks the policy to prioritize exploration as well as maximizing the returns but this improves the learning. So in non-standard Indian stock market conditions, SAC will potentially be able to balance between exploiting the working strategies as well as exploring novel trading strategies.

            \vspace{+5pt}
            \textbf{Maximum Entropy Objective:}

            \begin{equation}
            \mu^* =  \arg\max_{\mu} {E} \left[\sum_{i=0}^{\infty} \gamma^i \left(r(x_i, b_i) + \alpha H(\mu(\cdot|x_i))\right)\right]
            \label{equation:sac-entropy}
            \end{equation}

            Where $\mu^*$ is the optimal policy.
            $H(\mu(\cdot|x_i))$ represents the entropy of the policy at state $x_i$.
            $\alpha$ is the temperature parameter controlling the balance between maximizing rewards and maximizing entropy\cite{haarnoja2018softactorcriticoffpolicymaximumm}.

            \vspace{+5pt}
            \textbf{Soft Q-function:}

            \begin{equation}
            R^{\mu}(x_i, b_i) = r(x_i, b_i) + \gamma {E}_{x' \sim p} \left[ S^{\mu}(x_{i+1}) \right]
            \label{equation:sac-q}
            \end{equation}

            where $S^{\mu}(x_{i+1})$ is the soft value function, defined as:

            \begin{equation}
            S^{\mu}(x_{i+1}) =  {E}_{x' \sim p, b' \sim \mu} \left[R^{\mu}(x_{i+1}, b_{i+1}) - \alpha \log \mu(b_{i+1}|x_{i+1})\right]
            \label{equation:sac-v}
            \end{equation}

            The explicit working of SAC algorithm in current modelling of Indian financial marketing is explained in following steps:

            \begin{algorithm}[H]
            \caption{Soft \emph{Actor-Critic} Algorithm}
            \begin{algorithmic}[1]
                \State Initialize parameters: $\theta$ $\rightarrow$ \emph{actor} network ($\mu_{\theta}$), $\Phi_1$ $\rightarrow$ first \emph{critic} network ($Q_{\Phi_{1}}$), $\Phi_2$ $\rightarrow$ second \emph{critic} network ($Q_{\Phi_{2}}$)
                \State Initialize policy $\mu_{\theta}(x)$, Q-values networks $Q_{\phi_1}(x, b)$, $Q_{\phi_2}(x, b)$, target $Q$-networks \& temperature ($\alpha$)
                \State Initialize the replay buffer $D$
                \For {each iteration}
                    \For {each environment step}
                        \State Observe current state $x_t$ (technical indicators, inclination score)
                        \State Select action $b_t$ $\sim$ $\mu_{\theta}(b_t | x_t)$
                        \State Execute action $b_t$
                        \State Calculate reward $r_t$
                        \State Next State $x_{t+1} \sim P(x_{t+1} | x_t, b_t)$
                        \State Store in replay buffer $D \leftarrow D~\cup \{x_t, b_t, r_t(x_t, b_t), x_{t+1}\}$
                        \State Sample mini-batch from D \& update Q-values, policy \& $\alpha$
                        \State Soft update target $Q$-networks
                    \EndFor
                \EndFor
            \end{algorithmic}
            \end{algorithm}

        }
    }
}


\section{Experiments \& Results}
{
    In this section, before conducting the experiment we have decided upon the relevant evaluation metrics and methodologies to assess our results and findings.

    We have done it in three phases: A. \emph{Trading without sentiment}, B. \emph{Trading with partial sentiment infusion} and C. \emph{Trading with sentiment deep infusion}. It is structured this way to give a comprehensive choice to traders and investors to have options to choose from while defining their requirements and trading timeline. For experiment, we have allocated an initial portfolio of 1 million INR with trade timeline of approximately a year (\emph{June 2023} to \emph{June 2024}).

}

\subsection{Evaluation Metrics}
{
    To comprehensively evaluate the performance of our selected algorithms, we have shortlisted a few qualitative and quantitative evaluation metrics. These metrics are selected to explain the algorithm selection for different kinds of investment appetite and trading requirements i.e. aggressive return focus trading and risk averse trading.

    \subsubsection{Annual Returns}{
        This is a useful metric to evaluate long-term growth directly since it shows the annual overall profitability of our RL algorithms’ investment strategy.

        \begin{equation}
            Annual~Return = (\frac{P_A - P_0}{P_0} - 1) \times 100
            \label{equation:annual-returns}
        \end{equation}

        $P_A$ is annual value of portfolio and $P_0$ is the initial portfolio
    }
    \subsubsection{Sharpe Ratio}{
        This is an estimate of the risk-adjusted returns by factoring volatility. Higher values are preferred since this tells the returns per unit of risk.

        \begin{equation}
            Sharpe~Ratio = \frac{R_A - R_{ARF}}{\sigma_A} \quad \text{where $R_f$ is the risk-free rate.}
            \label{equation:sharpe-ratio}
        \end{equation}
    }
    \subsubsection{Stability}{
        This metric conveys the consistency of our algorithmic strategy in generating reliable profitable returns in various conditions of the market.

        \begin{equation}
            S_t = 1 - \frac{\sigma_{PG}}{\chi_{PG}}
            \label{equation:stability}
        \end{equation}

        $\sigma_{PG}$ is Standard deviation of portfolio growth and $\chi_{PG}$ is mean portfolio growth.
    }
    \subsubsection{Annual volatility}{
        This metric is useful to determine and manage the risk exposure by qualitatively measuring the price fluctuations. Low values mean lower risk.

        \begin{equation}
            \begin{split}
            Annual~Volatility &= \sigma_d \times \sqrt{252} \\
            &\quad \text{here $\sigma_d$ is daily return standard deviation.}
            \end{split}
            \label{equation:volatility}
        \end{equation}

    }
    \subsubsection{Max drawdown}{
        This is the delta between local maxima and local minima of a price fluctuation over a defined time period or in the equity curve. Lower max drawdown signifies lesser risk.

        \begin{equation}
            Max~Drawdown = \frac{P_H - P_L}{P_L}
            \label{equation:drawdown}
        \end{equation}

        Here $P_H$ and $P_L$ is the portfolio value at highest and lowest respectively.
    }
}

\subsection{Trading without Sentiment}
{
    Initially, the five selected DRL agents were traded on the proposed timeline without incorporating sentiment information. We used only technical indicators in our state space. The detailed agent-wise performances are as follows:


    \begin{table}[H]
    \centering
    \begin{tabular}{lrrrrr}
    \toprule
    Metric & A2C & DDPG & TD3 & PPO & SAC \\
    \midrule
    Annual return & 0.16 & 0.31 & 0.28 & 0.45 & 0.34 \\
    Annual volatility & 0.15 & 0.14 & 0.19 & 0.15 & 0.14 \\
    Sharpe ratio & 1.06 & 2.09 & 1.40 & 2.50 & 2.14 \\
    Stability & 0.58 & 0.83 & 0.66 & 0.91 & 0.80 \\
    Max drawdown & -0.13 & -0.09 & -0.15 & -0.06 & -0.11 \\
    \bottomrule
    \end{tabular}
    \vspace{+5pt}
    \caption{Performance Metrics of DRL Agents without Sentiment Infusion}
    \label{tab:table_without_sentiment}
    \end{table}

    \begin{itemize}
        \item \textbf{A2C}: It performed moderately as compared to other agents with lower returns and stability. It also performed poorly in drawdown (-0.1293) which is the maximum as compared to other agents. Drawdown indicates the maximum loss from a peak to a trough and its lower values are preferable.
        \item \textbf{DDPG}: It outperformed A2C in annual returns, and it showed the lowest volatility indicating more stability. It has a high Sharpe ratio which implies good risk-adjusted returns.
        \item \textbf{TD3}: It performed moderately on returns and showed a decent Sharpe ratio.
        \item \textbf{PPO}: Distinctively the highest annual return, stability, Sharpe ratio, and lowest max drawdown.
        \item \textbf{SAC}: Its Sharpe ratio is nearly as good as PPO and a strong return trailing PPO.
    \end{itemize}

    It was observed that PPO performed better than all other agents with the highest annual return and Sharpe ratio. This signifies that PPO consistently balanced risk and returns. For risk-averse investors, lower annual volatility is preferable, PPO showed moderate volatility at 0.1531, whereas DDPG has the lowest volatility at 0.1358.

    Thus, if the primary focus is the maximization of profits and risk-adjusted returns, PPO should be considered, since it showed the highest returns as well as the maximum Sharpe ratio. PPO also showed high stability which indicates its consistent performance and the lowest drawdown means that PPO also managed to keep the potential losses to a minimum.

    From the point of view of risk-averse trading, DDPG can be a better option, since it showed the least volatility and relatively low drawdown. Moreover, high stability and Sharpe ratio indicate that it can perform reliably by maintaining a good balance of annual returns and risks.

    From this experiment, we developed an understanding that in the absence of sentiment infusion the agents of both paradigms i.e. policy optimization based and Q-learning along with policy optimization, have paved the scope for improvement in performance if financial markets’ contextual awareness is available to our autonomous agents. We will be validating this idea in the next experiment.

}

\subsection{Trading with Partial Sentiment Infusion}
{
    We incorporated the sentiment inclination in the state space using the inclination score and observed the results. The individual agent-wise performance analysis is as follows:

        \begin{figure*}[t]
        \centering
        \begin{subfigure}[b]{0.32\textwidth}
            \includegraphics[width=\linewidth,
                alt={Line chart of A2C portfolio value from June 2023 to June 2024, without sentiment (blue) and with sentiment in the state space (green).}]{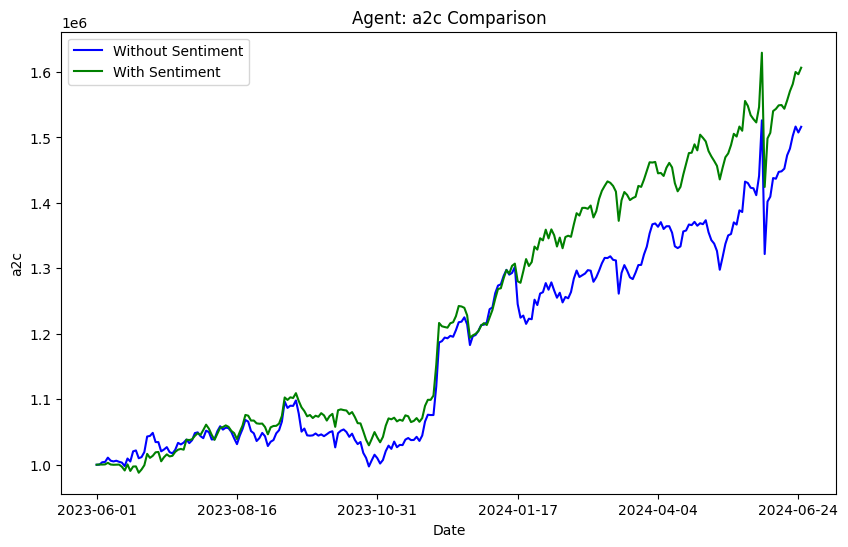}
            \caption{A2C}
            \label{fig:S_a2c}
        \end{subfigure}\hfill
        \begin{subfigure}[b]{0.32\textwidth}
            \includegraphics[width=\linewidth,
                alt={Line chart of DDPG portfolio value from June 2023 to June 2024, without sentiment (blue) and with sentiment in the state space (green).}]{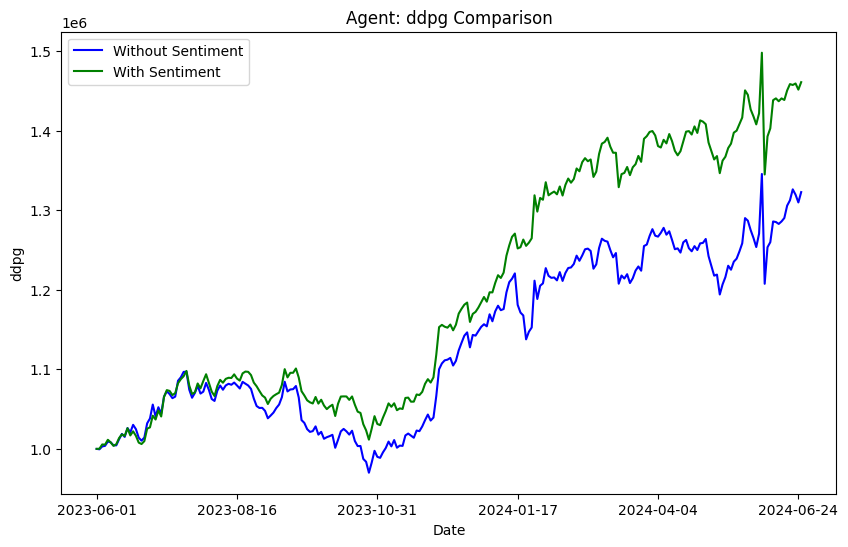}
            \caption{DDPG}
            \label{fig:S_ddpg}
        \end{subfigure}\hfill
        \begin{subfigure}[b]{0.32\textwidth}
            \includegraphics[width=\linewidth,
                alt={Line chart of TD3 portfolio value from June 2023 to June 2024, without sentiment (blue) and with sentiment in the state space (green).}]{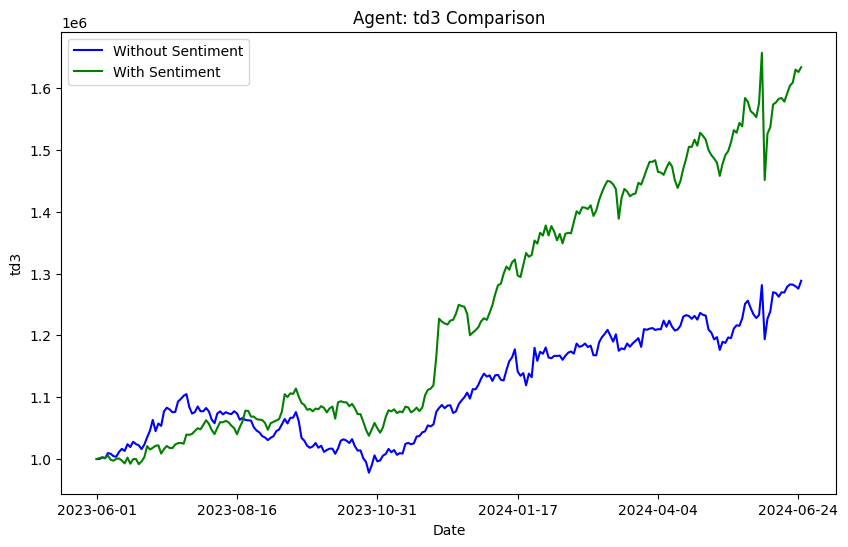}
            \caption{TD3}
            \label{fig:S_td3}
        \end{subfigure}

        \begin{subfigure}[b]{0.32\textwidth}
            \includegraphics[width=\linewidth,
                alt={Line chart of PPO portfolio value from June 2023 to June 2024, without sentiment (blue) and with sentiment in the state space (green).}]{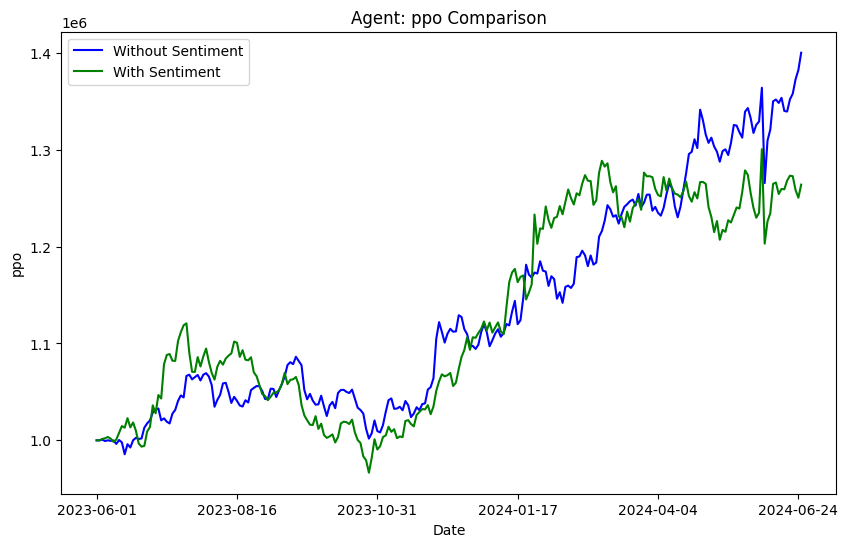}
            \caption{PPO}
            \label{fig:S_ppo}
        \end{subfigure}\hspace{0.02\textwidth}
        \begin{subfigure}[b]{0.32\textwidth}
            \includegraphics[width=\linewidth,
                alt={Line chart of SAC portfolio value from June 2023 to June 2024, without sentiment (blue) and with sentiment in the state space (green).}]{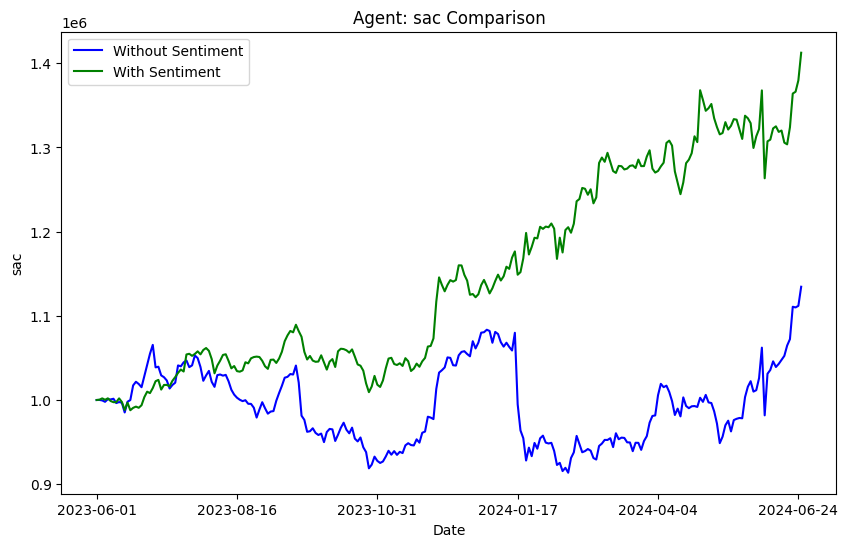}
            \caption{SAC}
            \label{fig:S_sac}
        \end{subfigure}
        \caption{Annual Return of DRL Agents with Partial Sentiment Infusion}
        \label{fig:annual_returns_all}
    \end{figure*}


    \begin{table}[h]
        \centering
        \begin{tabular}{lrrrrr}
        \toprule
        Metric & A2C & DDPG & TD3 & PPO & SAC \\
        \midrule
        Annual return & 0.61 & 0.46 & 0.63 & 0.26 & 0.41 \\
        Annual volatility & 0.21 & 0.19 & 0.21 & 0.19 & 0.18 \\
        Sharpe ratio & 2.36 & 2.15 & 2.46 & 1.32 & 2.01 \\
        Stability & 0.93 & 0.87 & 0.94 & 0.71 & 0.90 \\
        Max drawdown & -0.13 & -0.10 & -0.12 & -0.14 & -0.08 \\
        \bottomrule
        \end{tabular}
        \vspace{+5pt}
        \caption{Performance metrics for Sentiment Aware DRL}
        \label{tab:partial_sentiment}
    \end{table}

    \begin{itemize}
        \item \textbf{A2C}: It achieved one of the highest returns, thus making it the most profitable in this scenario. It also has a high Sharpe ratio, which implies that it provides good risk-adjusted returns.
        \item \textbf{DDPG}: It too showed improved returns and moderate volatility. It has a good sharp ratio and better stability, though less than A2C. It is also less risky owing to a relatively low drawdown.
        \item \textbf{TD3}: In terms of return, volatility, and max-drawdown it performed close to A2C. In terms of stability, it exhibited to be the most stable of all agents which is also demonstrated by its Sharpe ratio. With the high Sharpe ratio, it gives the best returns per unit of risk taken.
        \item \textbf{PPO}: Its returns dropped significantly, and it has a much lower Sharpe ratio, thus giving weaker risk-adjusted returns.
        \item \textbf{SAC}: It showed good returns, and a moderate volatility i.e. it’s less risky in terms of fluctuations while maintaining a strong Sharpe ratio and lowest drawdown.
    \end{itemize}

    We observed that the performance of A2C, DDPG, TD3, and SAC improved in terms of annual returns. PPO suffered a decline in its performance which can be attributed to its policy design. PPO tries to prevent large shifts in policy by penalizing large updates. The sentiment scores of real-time data may fluctuate which could cause PPO to be more conservative in updating its policy. Thus, in this scenario, PPO should be avoided.

    For maximizing profits TD3 is more suitable as it showed the highest performance in most of the metrics, indicating stability and consistency in terms of risk-adjusted returns. On the other hand, for risk-averse strategies, SAC is preferable since it provides a balance of low volatility, minimal drawdown, and consistent returns. The inclusion of sentiment-based contextual awareness of the market has upgraded the performance of agents based on the technique of Q-learning with policy optimization.

    This infusion-based improvement highlights the importance of further deeper infusion of contextual awareness into our model’s most dynamic components such as reward and state space combined. We will explore this in phase 3 of our experiment.


}

\subsection{Trading with Sentiment Deep Infusion}
{
    For qualitative as well as quantitative infusion of financial contextual awareness, we will be using the proposed method in section ~\ref{sec:reward} for reward-based infusion. We will be following the prior methodology of state-based infusion. Detailed agent-wise performance is as follows:

        \begin{figure*}[t]
        \centering
        \begin{subfigure}[b]{0.32\textwidth}
            \includegraphics[width=\linewidth,
                alt={Line chart of A2C portfolio value from June 2023 to June 2024, without sentiment (blue) and with sentiment in both the state space and the reward (green).}]{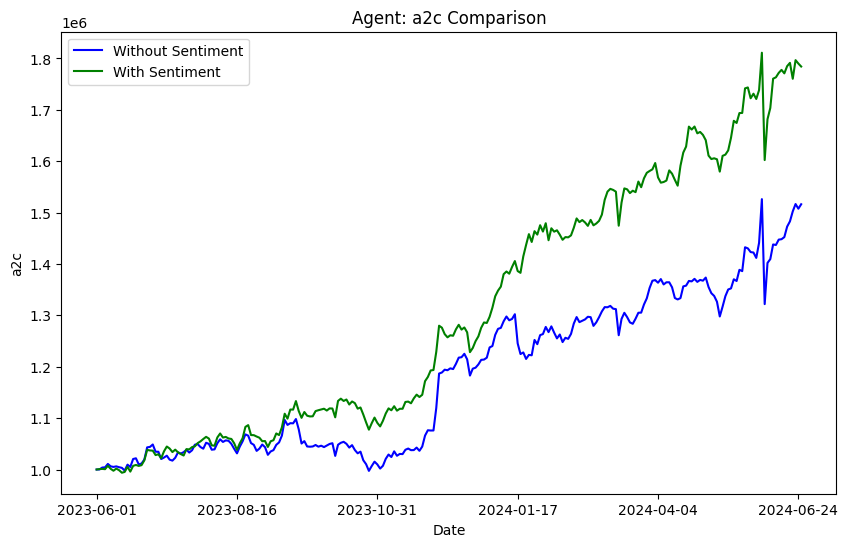}
            \caption{A2C}
            \label{fig:SR_a2c}
        \end{subfigure}\hfill
        \begin{subfigure}[b]{0.32\textwidth}
            \includegraphics[width=\linewidth,
                alt={Line chart of DDPG portfolio value from June 2023 to June 2024, without sentiment (blue) and with sentiment in both the state space and the reward (green).}]{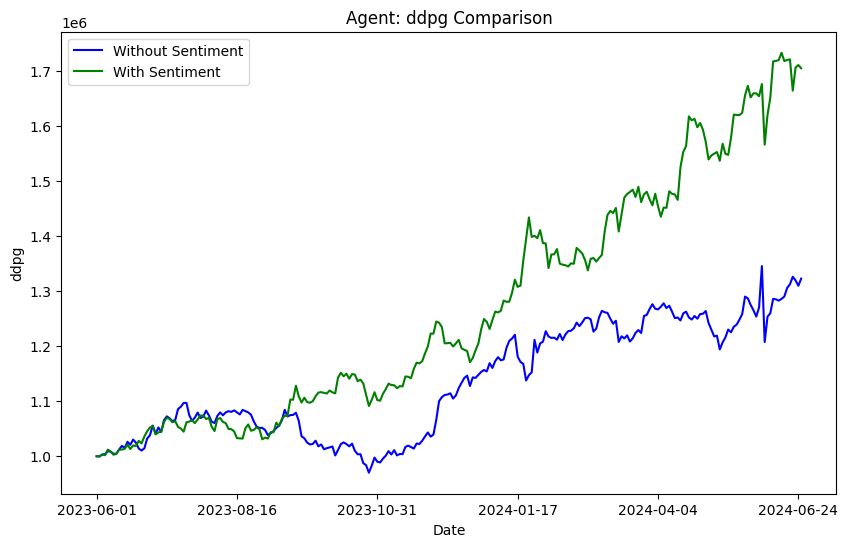}
            \caption{DDPG}
            \label{fig:SR_ddpg}
        \end{subfigure}\hfill
        \begin{subfigure}[b]{0.32\textwidth}
            \includegraphics[width=\linewidth,
                alt={Line chart of TD3 portfolio value from June 2023 to June 2024, without sentiment (blue) and with sentiment in both the state space and the reward (green).}]{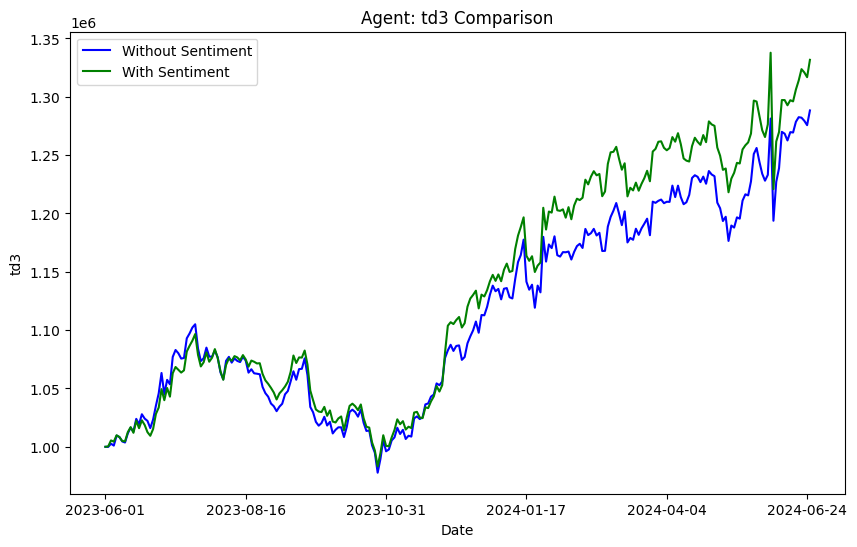}
            \caption{TD3}
            \label{fig:SR_td3}
        \end{subfigure}

        \begin{subfigure}[b]{0.32\textwidth}
            \includegraphics[width=\linewidth,
                alt={Line chart of PPO portfolio value from June 2023 to June 2024, without sentiment (blue) and with sentiment in both the state space and the reward (green).}]{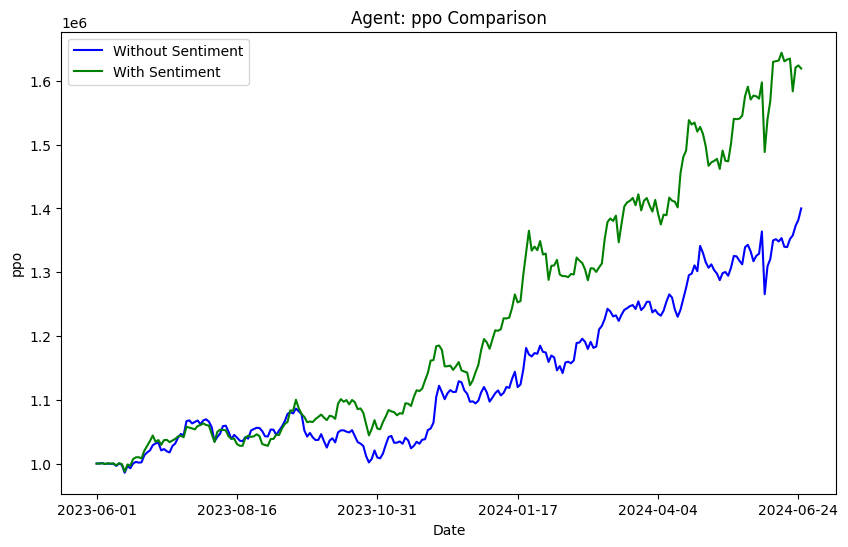}
            \caption{PPO}
            \label{fig:SR_ppo}
        \end{subfigure}\hspace{0.02\textwidth}
        \begin{subfigure}[b]{0.32\textwidth}
            \includegraphics[width=\linewidth,
                alt={Line chart of SAC portfolio value from June 2023 to June 2024, without sentiment (blue) and with sentiment in both the state space and the reward (green).}]{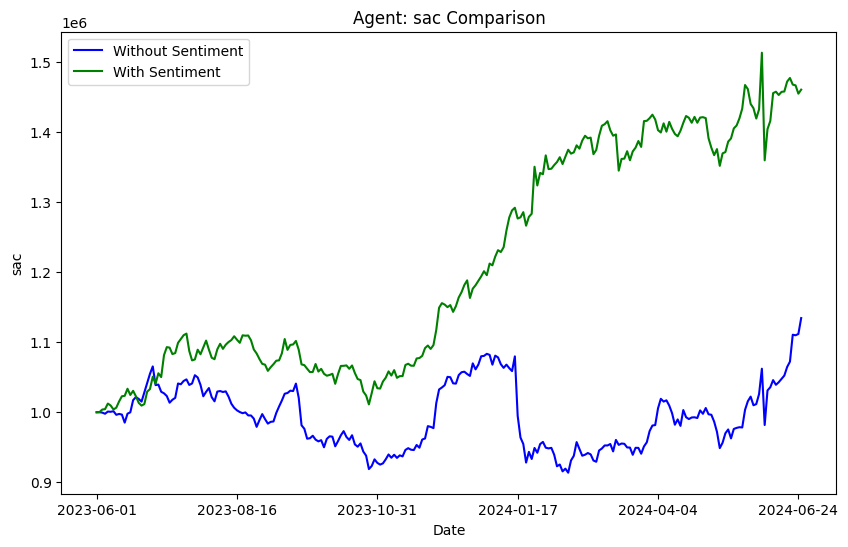}
            \caption{SAC}
            \label{fig:SR_sac}
        \end{subfigure}
        \caption{Annual Return of DRL Agents with Deep Sentiment Infusion}
        \label{fig:RL-with-state-reward}
    \end{figure*}

    \begin{table}[H]
        \centering
        \begin{tabular}{lrrrrr}
        \toprule
        Metric & A2C & DDPG & TD3 & PPO & SAC \\
        \midrule
        Annual return & 0.26 & 0.70 & 0.33 & 0.25 & 0.46 \\
        Annual volatility & 0.21 & 0.20 & 0.17 & 0.19 & 0.20 \\
        Sharpe ratio & 1.23 & 2.75 & 1.78 & 1.24 & 2.02 \\
        Stability & 0.68 & 0.95 & 0.81 & 0.66 & 0.86 \\
        Max drawdown & -0.15 & -0.07 & -0.10 & -0.15 & -0.10 \\
        \bottomrule
        \end{tabular}
        \vspace{+5pt}
        \caption{Performance metrics of DRL Agents with Deep Sentiment Infusion}
        \label{tab:with_sentiment_2}
        \end{table}

    \begin{itemize}
        \item \textbf{A2C}: Showed moderate profitability at an annual return of about ~26\%. Its Sharpe ratio of 1.23 implies acceptable risk-adjusted returns.
        \item \textbf{DDPG}: Showed the highest annual return and highest Sharpe ratio. High stability and low max drawdown make it more consistent and profitable.
        \item \textbf{TD3}: Its annual return dropped and the Sharpe ratio declined, although its max drawdown improved slightly.
        \item \textbf{PPO}: PPO struggled to generate significant profit as compared to prior, in other metrics it performed closely to A2C.
        \item \textbf{SAC}: Showed a decent level of returns. A Sharpe ratio of 2.02 and high stability imply it is capable of good risk-adjusted returns.
    \end{itemize}

    We observed that the performance of A2C and PPO in this case remained consistent with partial sentiment infusion. SAC showed improvement in annual returns and the Sharpe ratio. Whereas, the performance of DDPG showed significant improvement with higher returns, better risk-adjusted performance, and smaller drawdown.
    TD3 experienced a decrease in performance by becoming less profitable, indicating that it could be more sensitive to how the infusion of sentiment impacted the reward structure.

    The agents that optimize the policy as well as the Q-value (DDPG, SAC), become more stable upon deeper sentiment infusion except for TD3. Although TD3 is an improved version of DDPG but DDPG consistently showed improved performance upon deeper sentiment infusion whereas TD3 produced poor returns. This behavior arises from the more complex nature of TD3 to handle updates. TD3 uses double Q-learning to ensure conservative value updates. The infusion of sentiment might cause increased variability in the reward signal which caused TD3 to become overly conservative and under explore leading to suboptimal policies and hence drop in performance.


}


\section{Conclusions}
{
    In this paper, we simulated the real-world trading of curated stocks via reinforcement learning agents that interacted with the dynamically changing market condition modeled as a reinforcement learning framework. We selected the \emph{actor-critic-based} agents with distinct policy design specifically, \emph{A2C} and \emph{PPO} are based only on policy optimization, whereas \emph{DDPG, TD3} \& \emph{SAC} are based on policy optimization combined with Q-learning principles.

    We initially performed the trading operation without any qualitative features using only technical indicators. Introducing partial sentiment infusion improved the performance of all agents except PPO. Deeper sentiment infusion led to consistently improved performance of policy optimization and Q-learning-based agents (DDPG and SAC) while TD3 due to a more conservative approach to value estimation demonstrated decreased performance. Notably, DDPG consistently exhibited performance improvement with SAC also showing significant improvement and ranking just below DDPG.

    For further research, \emph{Explainable Reinforcement Learning} (XRL) techniques\cite{Milani20241,Amitai2024} are recommended to demonstrate the impact of various features such as sentiment score and various technical indicators on the agent’s decision-making process.
}


\section*{Disclosure of Interests}
The authors have no competing interests to declare that are relevant to the content of this research.

\bibliographystyle{ACM-Reference-Format}
\bibliography{reference}

\end{document}